\documentclass[lettersize,journal]{IEEEtran}
\usepackage{amsmath,amsfonts}
\usepackage{algorithmic}
\usepackage{array}
\usepackage{multirow} 
\usepackage{booktabs} 
\usepackage{tabularx}
\usepackage{amsmath}  
\usepackage[caption=false,font=normalsize,labelfont=sf,textfont=sf]{subfig}
\usepackage{textcomp}
\usepackage{stfloats}
\usepackage{url}
\usepackage{verbatim}
\usepackage{graphicx}
\usepackage{float} 
\usepackage{booktabs} 
\usepackage{tabularx} 
\usepackage{multirow}
\newcolumntype{C}{>{\centering\arraybackslash}X} 
\usepackage{verbatim}
\usepackage{array} 
\usepackage{colortbl}
\usepackage{xcolor} 
\usepackage{amsmath}
\newcolumntype{Y}{>{\centering\arraybackslash}X}
\usepackage{pifont}   
\newcommand{\cmark}{\ding{51}}

\def\BibTeX{{\rm B\kern-.05em{\sc i\kern-.025em b}\kern-.08em
    T\kern-.1667em\lower.7ex\hbox{E}\kern-.125emX}}
\usepackage{balance}
\begin{document}
\title{Optimization of Prompt Learning via Multi-Knowledge Representation for Vision-Language Models}
\author{Enming Zhang, Bingke Zhu, Yingying Chen, Qinghai Miao, Ming Tang and Jinqiao Wang
\thanks{
This work was supported by the National Natural Science Foundation of China (62271485) and the Beijing Municipal Science and Technology Project (Z231100007423004). (Corresponding author: Qinghai Miao.)

Enming Zhang and Qinghai Miao are with the School of Artificial Intelligence, University of Chinese Academy of Sciences, Beijing 100049, China (e-mail: zhangenming23@mails.ucas.ac.cn; miaoqh@ucas.ac.cn).

Bingke Zhu, Yingying Chen and Ming Tang are with the Foundation Model Research Center, Institute of Automation, Chinese Academy of Sciences, Beijing 100190, China, and with Wuhan AI Research, Wuhan 430073, China (e-mail: bingke.zhu@nlpr.ia.ac.cn; yingying.chen@nlpr.ia.ac.cn; tangm@nlpr.ia.ac.cn).

Jinqiao Wang is with the Foundation Model Research Center, Institute of Automation, Chinese Academy of Sciences, Beijing 100190, China, also with the School of Artificial Intelligence, University of Chinese Academy of Sciences, Beijing 100049, China, also with Wuhan AI Research, Wuhan 430073, China, and with Peng Cheng Laboratory, Shenzhen 518066, China (e-mail: jqwang@nlpr.ia.ac.cn).}
}

\markboth{Journal of \LaTeX\ Class Files,~Vol.~18, No.~9, September~2020}%
{How to Use the IEEEtran \LaTeX \ Templates}

\maketitle

\begin{abstract}
Vision-language models (VLMs), such as CLIP, play a foundational role in various cross-modal applications. To fully leverage the potential of VLMs in adapting to downstream tasks, context optimization methods such as prompt tuning are essential. However, one key limitation is the lack of diversity in prompt templates, whether they are hand-crafted or learned through additional modules. This limitation restricts the capabilities of pretrained VLMs and can result in incorrect predictions in downstream tasks. To address this challenge, we propose context optimization with multi-knowledge representation (CoKnow), a framework that enhances prompt learning for VLMs with rich contextual knowledge. To facilitate CoKnow during inference, we train lightweight semantic knowledge mappers, which are capable of generating multi-knowledge representations for an input image without requiring additional priors. Experimentally, we conduct extensive experiments on 11 publicly available datasets, demonstrating that CoKnow outperforms a series of previous methods.
\end{abstract}


\section{Introduction}
\begin{figure}[t]
\centering
\includegraphics[width=3.6in]{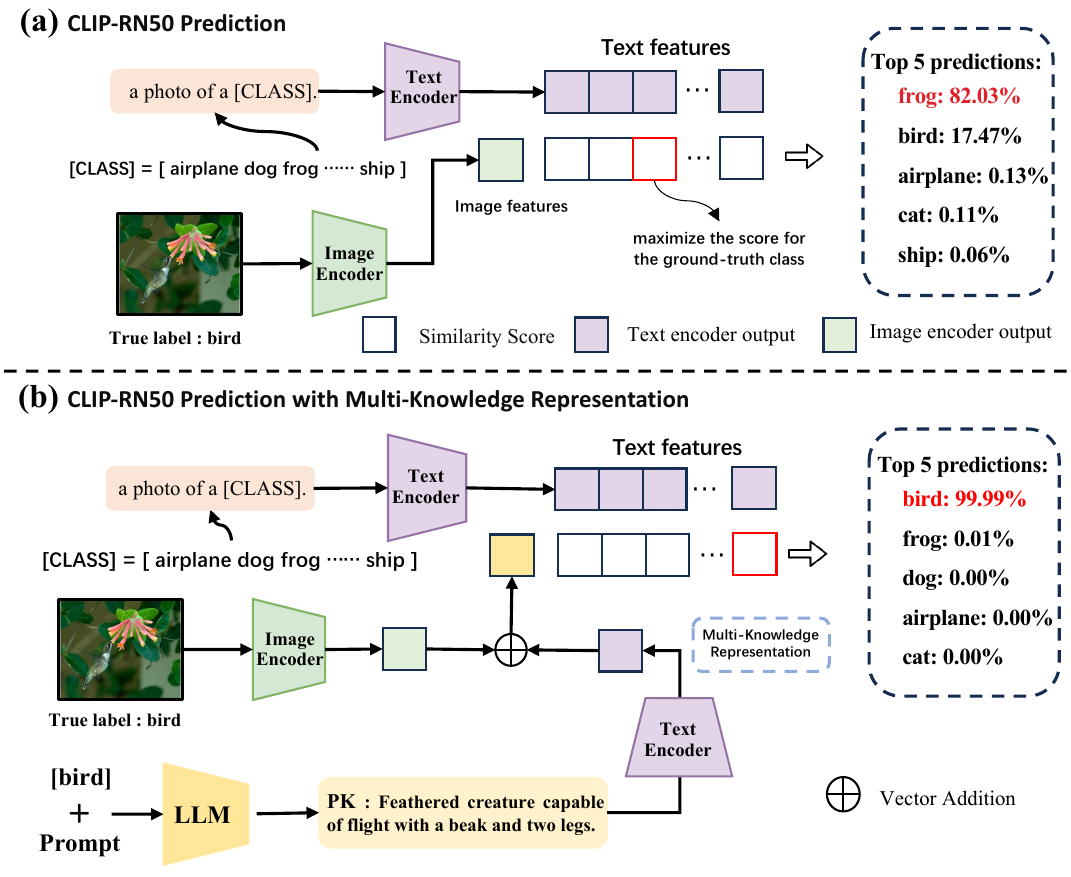}
    \caption{\textbf{The result predicted by CLIP without introducing multi-knowledge representation is shown in Figure (a). The result predicted by CLIP after introducing multi-knowledge representation is shown in Figure (b). Introducing multi-knowledge representation improves the prediction accuracy.}}
\label{fig:1}
\end{figure}

\IEEEPARstart{B}\ y learning representations of millions of image-text pairs in a shared embedding space, large-scale pretrained vision-language models (VLMs), \textit{e.g.,} CLIP \cite{clip}, have become powerful tools for a wide range of multimodal applications \cite{tmm1,tmm2,tmm3,tmm4,tmm5}.
The performance of VLMs is crucial for downstream tasks or applications that rely on VLMs as core components. However, plain VLMs may not perform optimally in some fundamental tasks, leading to unmet expectations. Figure \ref{fig:1}(a) illustrates the classification task using CLIP on the CIFAR-10 \cite{Cifar-10} dataset. In this task, a specific template is used for text input, such as \textit{a photo of a [CLASS]}, where \textit{[CLASS]} is replaced with all categories. The text encoder generates the embedding for each category, whereas the image encoder produces the image embedding. These embeddings are then compared pairwise, with the category exhibiting the highest similarity to the image embedding being chosen as the final classification. For the case shown in Figure \ref{fig:1}(a), CLIP incorrectly identifies an image of a bird as a frog, scoring 82$\%$ for this misclassification.

To enhance the performance of CLIP \cite{clip} in downstream tasks, many methods have been proposed, which can be roughly categorized into two main types: fine tuning and prompt learning. To avoid the high computational load of full parameter tuning, fine-tuning methods typically freeze the CLIP parameters and introduce trainable additional modules to adapt to downstream tasks. Prompt learning methods, on the other hand, attempt to learn better contextual templates to replace manually crafted prompts such as \textit{a photo of a [CLASS]}.
These methods effectively improve the performance of CLIP, but there is still much room for improvement in many tasks, especially in few-shot tasks in low-resource scenarios.

The improvement space for pretrained VLMs in downstream tasks can come from the potential they have yet to be elicited. One possible explanation is that a simple text prompt may not adequately represent the complexity of an image, which typically contains more information than its captions do. In reality, a single object can be described in various ways, depending on the level of detail or perspective. While models such as CLIP have been trained on a vast dataset of image-text pairs and have learned rich relationships between visual and textual content, they may struggle with simple text prompts that fail to capture the full semantic meaning of an image. To fully utilize the capabilities of CLIP, we propose enhancing the prompt context by incorporating knowledge from multiple perspectives at multiple abstraction levels, or in short, Multi-Knowledge. In downstream tasks of image recognition, we define the global representation of a piece of text in VLMs as a multi-knowledge representation. Specifically, a natural language description consists of multiple words, each represented by a token in VLMs. Each word also represents different pieces of knowledge, and their encoded global representations constitute the multi-knowledge representation.

\begin{figure*}[t]
    \centering
    \includegraphics[width=1.00\textwidth]{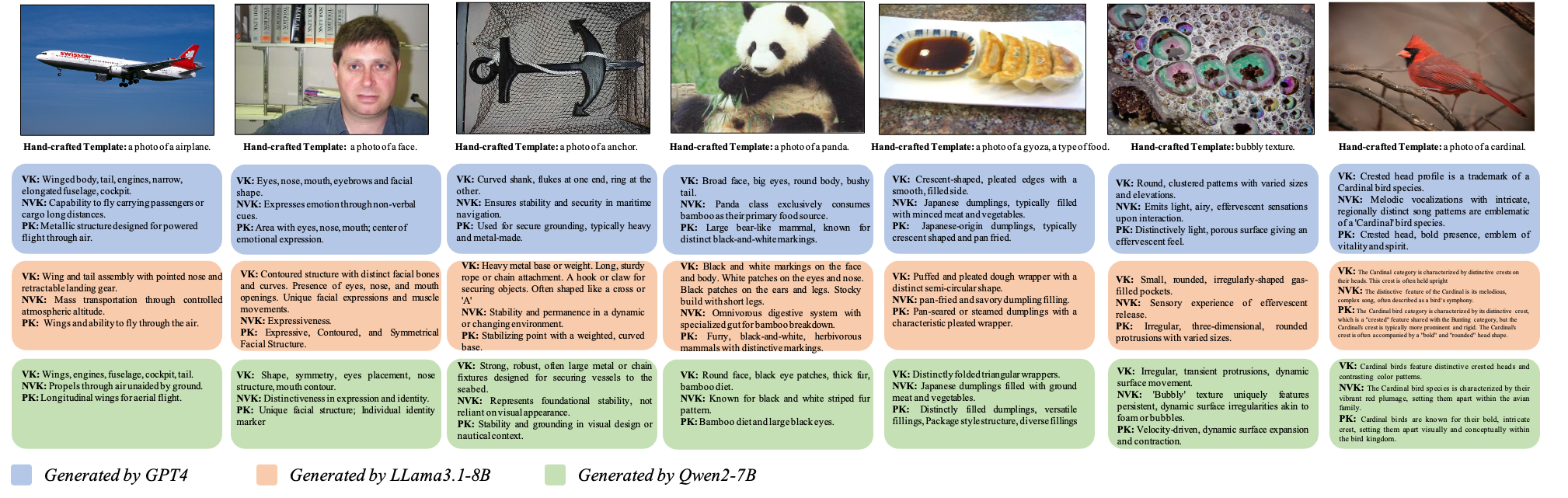}
    \caption{
\textbf{Instances of Multi-Knowledge. These include categories such as airplanes, human faces, ship wheels, pandas, gyoza food, and bubble textures. Here, blue represents the content generated by GPT-4, orange represents the content generated by LLama3.1, and green represents the content generated by Qwen2.}
}
    \label{figexam2}
\end{figure*}

The Multi-Knowledge method introduced in this study comprises three types. In Figure \ref{figexam2}, we present examples of Multi-Knowledge. The first is visual knowledge (VK), which includes captions describing the image or its category. For the example in Figure \ref{fig:1}, the visual knowledge for the given image could be \textit{Feathers, beak, wings, two-legged, small body size.} Second, and of greater importance, we introduce nonvisual knowledge (NVK) at a more abstract level beyond purely visual aspects. The NVK for the image in Figure \ref{fig:1} could be \textit{Migratory creatures mastering flight and song.} Third, we can combine multilevel descriptions, such as both VK and NVK, into a more comprehensive description called panoramic knowledge (PK). As shown in Figure \ref{fig:1}(b), the PK could be \textit{Feathered creatures capable of flight with a beak and two legs.} With the introduction of PK, which is encoded and added to the image embedding, the prediction for the input image, previously predicted as \textit{frog} in Figure \ref{fig:1}(a), is now predicted as the correct category \textit{bird} with 99.99$\%$ confidence. Furthermore, experimental results across the entire CIFAR-10 test set of 10,000 images show a 7.64$\%$ increase in prediction accuracy simply when PK is introduced. \textit{(The specific experimental results can be found in Section 3. )} This study demonstrates the essential role of multi-knowledge representation in enhancing downstream tasks.

To leverage multi-knowledge representation and the successful techniques of fine-tuning and soft prompt tuning for VLMs, we propose CoKnow, which consists of two learnable key modules, as shown in Figure \ref{coknow}.
The first module is an optimizer for prompt learning guided by multi-knowledge representation, allowing for adaptive learning of prompt templates rich in domain knowledge for various downstream tasks. The second module is a lightweight semantic knowledge mapper, which, once trained, can directly generate the corresponding multi-knowledge representation from images without additional input. This mapper is designed as a general module that can operate in a plug-and-play style with general VLMs and is not limited to CLIP. Notably, CoKnow leverages the large language model GPT-4 
\cite{gpt4} as the knowledge generator, automatically producing multi-knowledge representations via a set of simple prompt templates. Current large language models have demonstrated remarkable emergent intelligence, generalization capabilities, and knowledge processing abilities \cite{tmm6}. Experimentally, we conducted experiments on 11 publicly available datasets spanning various downstream domains via CoKnow. The results, which surpass several previous few-shot methods, confirm that CoKnow is an effective approach for VLM applications. Our main contributions are as follows:

\begin{itemize}
    {
    \item 
    To increase the performance of VLMs in downstream tasks such as classification, we propose multi-knowledge representations, including image captions at the visual level, nonvisual descriptions at the abstract level, and their combinations. Simultaneously, we validate the effectiveness of the multi-knowledge representation.

\item
    We introduce the CoKnow framework, which features a trainable prompt optimizer and lightweight visual-text semantic mappers, to support multi-knowledge representation in tasks based on vision-language models.

\item
    We provide effective and scalable methods for generating knowledge across diverse categories. For 11 publicly available datasets, we construct 6603 distinct knowledge descriptions corresponding to 2201 object categories via GPT-4. All resources will be made open source to support future research efforts.
    }
\end{itemize}

\begin{table*}[t]
    \centering
    \caption{\textbf{Comparison of image recognition methods from different series for vision-language models. "\cmark" indicates that this part of the content is utilized in the method.}}
    \footnotesize 
    \setlength{\tabcolsep}{4pt} 
    \renewcommand{\arraystretch}{1.2}
    \begin{tabularx}{\textwidth}{l|c|X|X|X|X} 
        \toprule
        \quad \textbf{Methods} & \textbf{Extra Modules} & \textbf{\quad \quad \quad  Soft Prompt} & \textbf{\quad \quad \quad Image Rep.} & \textbf{\quad \quad  Fixed Template Rep.} & \textbf{\quad  Multi-Knowledge Rep.} \\
        \midrule
        CLIP-Adapter \cite{gao2024clipada} & \cmark & \quad \quad \quad \quad \quad  - & \quad \quad \quad \quad \quad  \cmark & \quad \quad \quad \quad \quad \quad  \cmark & \quad \quad \quad \quad \quad \quad- \\
    
        Tip-Adapter \cite{Tip-adapter} & \cmark & \quad \quad \quad \quad \quad   - & \quad \quad \quad \quad \quad  \cmark & \quad \quad \quad \quad \quad \quad \cmark & \quad \quad \quad \quad \quad \quad- \\
     
        CLIP-A-self \cite{clip-gpt4} & \cmark & \quad \quad \quad \quad \quad   - & \quad \quad \quad \quad \quad  \cmark & \quad \quad \quad \quad \quad \quad - & \quad \quad \quad \quad \quad \quad \cmark \\
       
        CoOp \cite{coop} & - & \quad \quad \quad \quad \quad  \cmark & \quad \quad \quad \quad \quad  \cmark & \quad \quad \quad \quad \quad \quad  - & \quad \quad \quad \quad \quad \quad- \\
       
        CoCoOp \cite{cocoop} & \cmark & \quad \quad \quad \quad \quad  \cmark & \quad \quad \quad \quad \quad  \cmark & \quad \quad \quad \quad \quad \quad  - & \quad \quad \quad \quad \quad \quad- \\
        
        kgCoOp \cite{kg} & \cmark & \quad \quad \quad \quad \quad  \cmark & \quad \quad \quad \quad \quad  \cmark & \quad \quad \quad \quad \quad \quad \cmark & \quad \quad \quad \quad \quad \quad- \\
       
        \textbf{CoKnow(Ours)} & \cmark & \quad \quad \quad \quad \quad  \cmark & \quad \quad \quad \quad \quad  \cmark & \quad \quad \quad \quad \quad \quad \cmark & \quad \quad \quad \quad \quad \quad \cmark \\
        \bottomrule
    \end{tabularx}
    \label{tab:comparison_method}
\end{table*}

\section{Related work}
Currently, significant advancements are being made in vision-language models \cite{jia2021scaling,radford2021learning,singh2022flava,yuan2021florence,zhai2022lit}, which create rich multimodal representations between natural language and vision, facilitating a wide array of downstream tasks. Among these models, CLIP \cite{clip} stands out as a prominent method. By undergoing self-supervised training on 400 million image-text pairs, CLIP learns joint representations for images and text, showing outstanding performance across various downstream visual tasks.
Although pretrained VLMs can represent image-text pairs, applying them effectively to downstream tasks still faces several challenges. Recently, many studies have shown improved performance in downstream tasks, such as few-shot image recognition \cite{coop,clip-gpt4,Tip-adapter}, by customizing CLIP through fine tuning and prompt learning.

\subsection{Fine Tuning}

In downstream tasks with few samples, full-tuning \cite{over,mao2023context,zhang2022glipv2} of CLIP often leads to overfitting issues, resulting in poor generalization performance on test data. Recently, introducing additional modules to CLIP and fine-tuning them has proven to be an effective method for adapting it to downstream image recognition tasks. CLIP adopts linear probing \cite{clip,chen2020simple}, which involves introducing an additional linear layer for fine-tuning downstream tasks. CALIP \cite{guo2023calip} invokes a parameter-free attention mechanism to promote interactions between visual and textual modalities, exploring essential cross-modal informational features. Its parametric iteration, CALIP-F \cite{guo2023calip} particularly excelled in few-shot testing environments, showcasing outstanding efficiency. WiSE-FT \cite{WiSE-FT} proposes a weight-space ensemble approach that merges CLIP's pretrained and fine-tuned weights to bolster robustness against atypical distributions. Houlsby and colleagues \cite{houlsby2019parameter} introduced an adapter strategy that embeds learnable linear layers into transformer layers while immobilizing the network's backbone, providing a streamlined path for fine-tuning downstream tasks. Gao et al. \cite{gao2024clipada} formulated the CLIP-Adapter, which integrates feature adapters for nuanced fine-tuning within the visual or language branches. Zhang et al. \cite{Tip-adapter} launched Tip-Adapter, which employs a key-value cache model conceived from a limited-sample training set, enhancing initialization for quicker fine-tuning convergence. Lin et al. \cite{Cross-modal} ventured into cross-modal few-shot learning by repurposing textual features as training samples during the fine-tuning phase, paving a novel pathway for utilizing textual data in training. These methods adapt CLIP for downstream tasks by introducing additional modules. Similarly, we trained the semantic knowledge mappers, which can automatically generate corresponding multi-knowledge representations without the need for additional input, offering the advantage of plug-and-play functionality.

\subsection{Prompt Learning}
Another major challenge of applying large-scale pretrained VLMs such as CLIP to image recognition tasks is prompt engineering, which demands domain-specific expertise downstream and is extremely time-consuming, requiring substantial time investment for adjustments. To address this issue, CoOp \cite{coop} first introduced prompt learning into CLIP, achieving outstanding results in downstream few-shot image recognition tasks. Subsequent improvements have enabled a variety of tasks, including fine-grained object retrieval \cite{wangobj} and image-text classification \cite{guo2023texts}. Furthermore, CoCoOp \cite{cocoop} enhances generalization by incorporating image features into each prompt embedding via lightweight networks. ProGrad \cite{ProGrad} refines downstream task adaptation by correcting feedback gradients. Recently, KgCoOp \cite{kg} proposed constraining learnable prompt embeddings with common knowledge to enhance transfer generalization. These methods optimize learnable templates by leveraging image representations, and further, KgCoOp introduces handcrafted template representations for optimizing learnable templates.
\subsection{Multi-Knowledge Representation}
\begin{figure}[t]
    \centering
    \hspace{-1.1cm} 
    \includegraphics[width=3.5in]{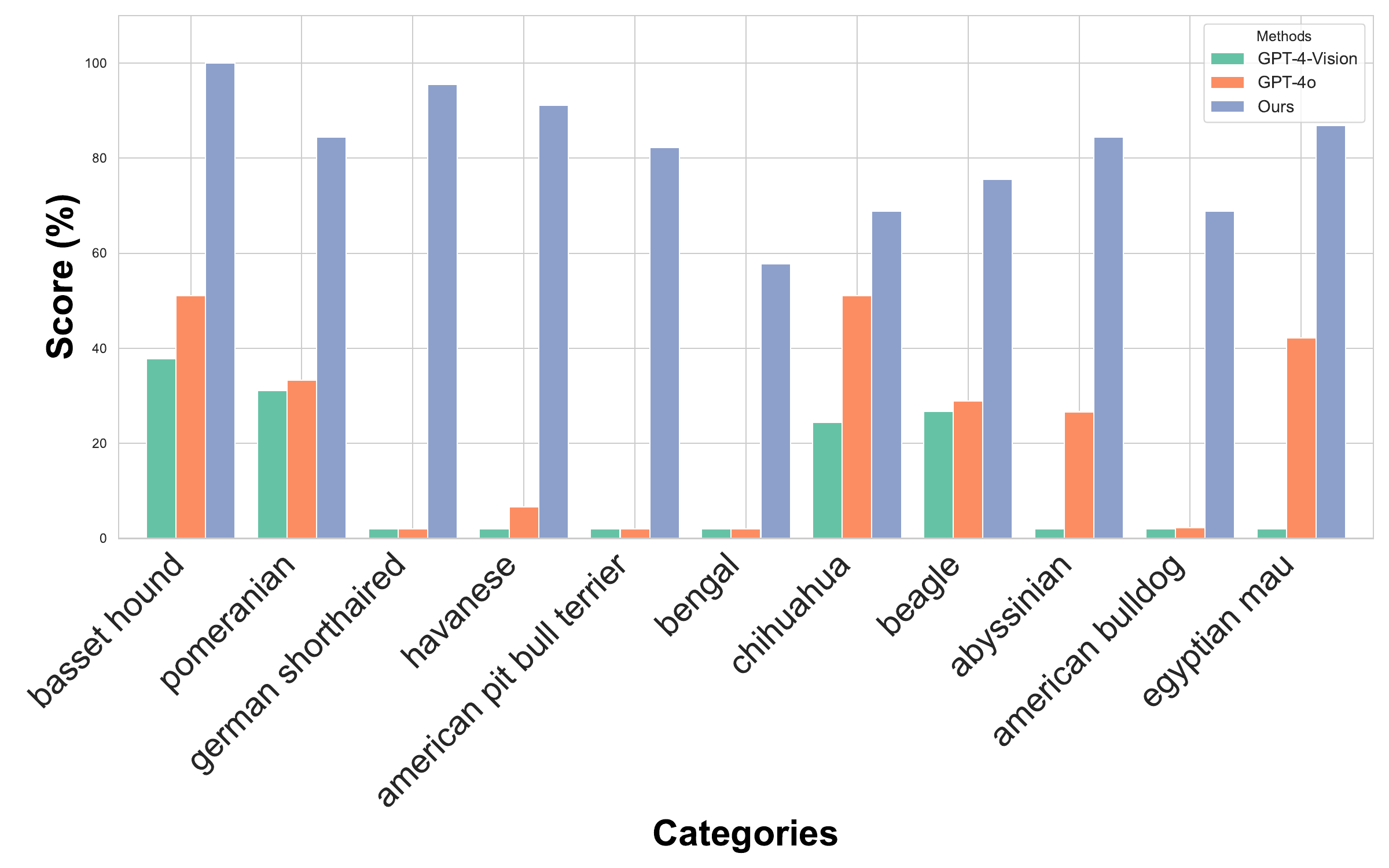}
    \caption{
    \textbf{Image classification is performed on the Oxford Pets dataset using multimodal large models and our method.}}
    
    \label{gpt4c}
\end{figure}

Both existing methods, prompt learning and model fine-tuning, overlook the potential for incorporating a broader range of domain-specific knowledge representations corresponding to image representations, which can be articulated via natural language. We refer to this global semantic representation described in natural language as multi-knowledge representation. Recently, CLIP-A-self \cite{clip-gpt4} leverages GPT-4 to generate multiple visual description texts as inputs to the text encoder and uses self-attention mechanism networks to aggregate them on each sentence, achieving good generalization effects. These visual descriptions are remarkably similar to the VK type within Multi-Knowledge. We first propose the use of multi-knowledge representation to optimize learnable contextual templates, delivering domain knowledge information to downstream tasks through adaptive means while reducing human involvement, thus making the process more convenient. We explored the effectiveness of other types of Multi-Knowledge, including VK, NVK, and PK. It is worth noting that we choose the powerful GPT-4 large language model version to generate multi-knowledge to assist our classification task, rather than directly using its multimodal capabilities for image classification. This decision is based on the fact that the multimodal version of GPT-4 still has significant limitations in fine-grained image question-answering tasks, with poor classification performance between fine-grained categories. Additionally, as a VQA model, it heavily relies on manually adjusted prompt settings, leading to inefficient classification and inevitable hallucination problems. In contrast, when GPT-4 has a clear understanding of a particular object category, i.e., when it clearly knows the name of the object, it can provide rich knowledge related to that object. Finally, we present the experimental results of using the GPT-4 Vision version and GPT-4o for direct image classification in Figure \ref{gpt4c}. These results are compared with those obtained by introducing PK on CLIP-RN50 (\textit{Ours}). The prompt we use is: \textit{You are an image classification AI assistant, please perform category recognition on this image.} The classification accuracy is below 30\% for most categories.

\begin{figure*}[t]
    \centering
    \includegraphics[width=1.00\textwidth]{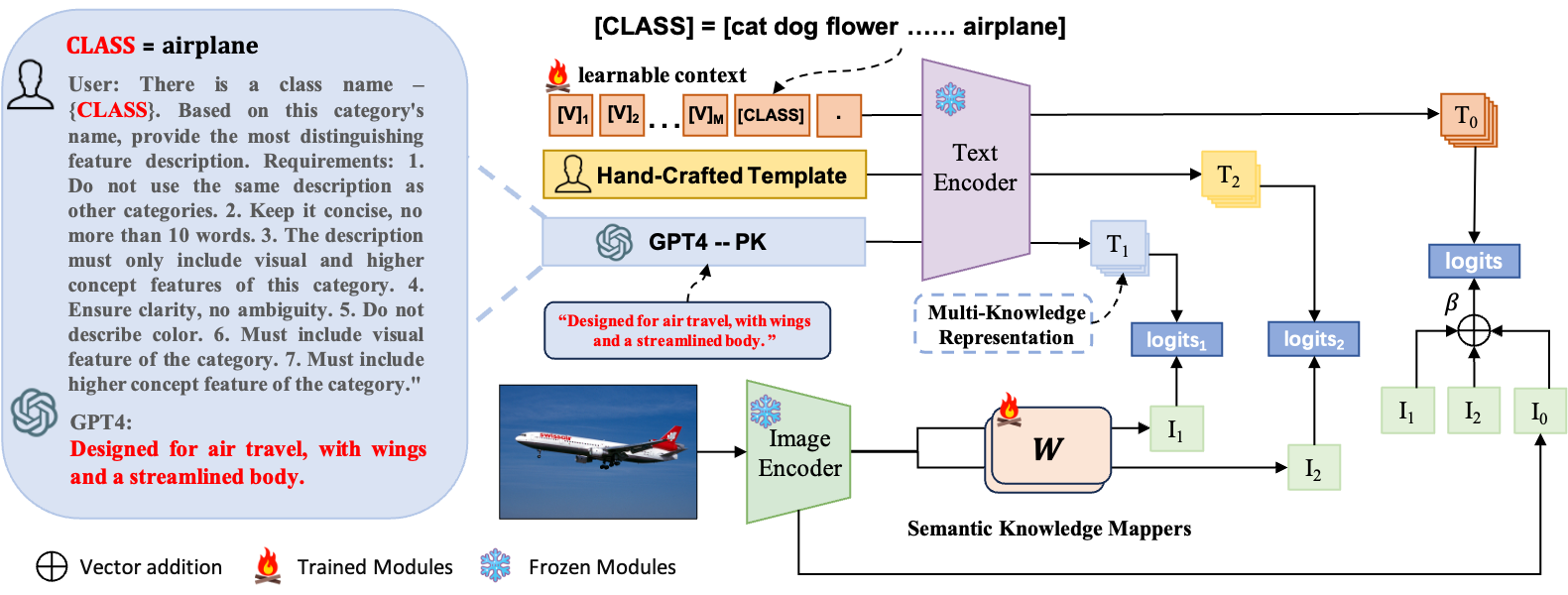}
    \caption{
\textbf{The overall architecture and training process of CoKnow. The figure illustrates the input of three types of templates into the text encoder: learnable context, Multi-Knowledge, and hand-crafted template, when the category is \textit{airplane}. The output of the image encoder undergoes contrastive loss calculations with the corresponding target template embeddings after passing through the semantic knowledge mappers. The original image representation and the mapped image representation are added with $\beta$ weights before performing contrastive loss calculations with the final embedding representation of the learnable contexts. Throughout the entire process, we train only lightweight semantic knowledge mappers to optimize context learning.}
}
    \label{coknow}
\end{figure*}

\begin{figure*}
\centering
\includegraphics[width=0.80\textwidth]{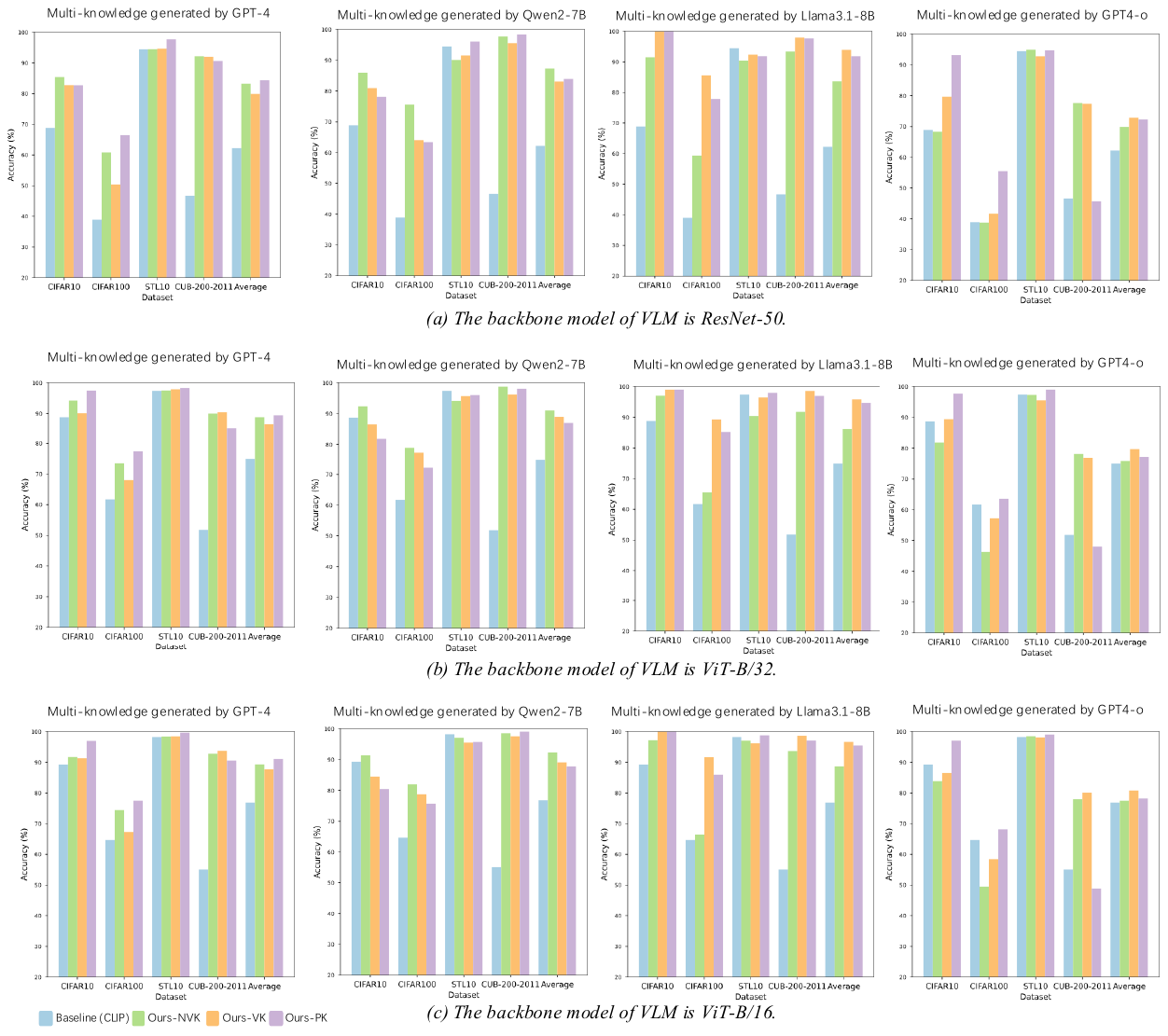}
    \caption{\textbf{The prediction results with the introduction of different Multi-Knowledge. (a) The figure presents the prediction results obtained via GPT-4, Qwen2, Llama3.1 and GPT4-o with a ResNet-50 backbone. (b) The figure illustrates the prediction results achieved with GPT-4, Qwen2, Llama3.1 and GPT4-o utilizing a ViT-B/32 backbone. (c) The figure displays the prediction results derived from GPT-4, Qwen2, Llama3.1 and GPT4-o employing a ViT-B/16 backbone. Among all the results, the Multi-Knowledge approach utilizing GPT-4 results in greater stability.}}
\label{zero_pre}
\end{figure*}

\begin{figure*}
\centering
\includegraphics[width=0.90\textwidth]{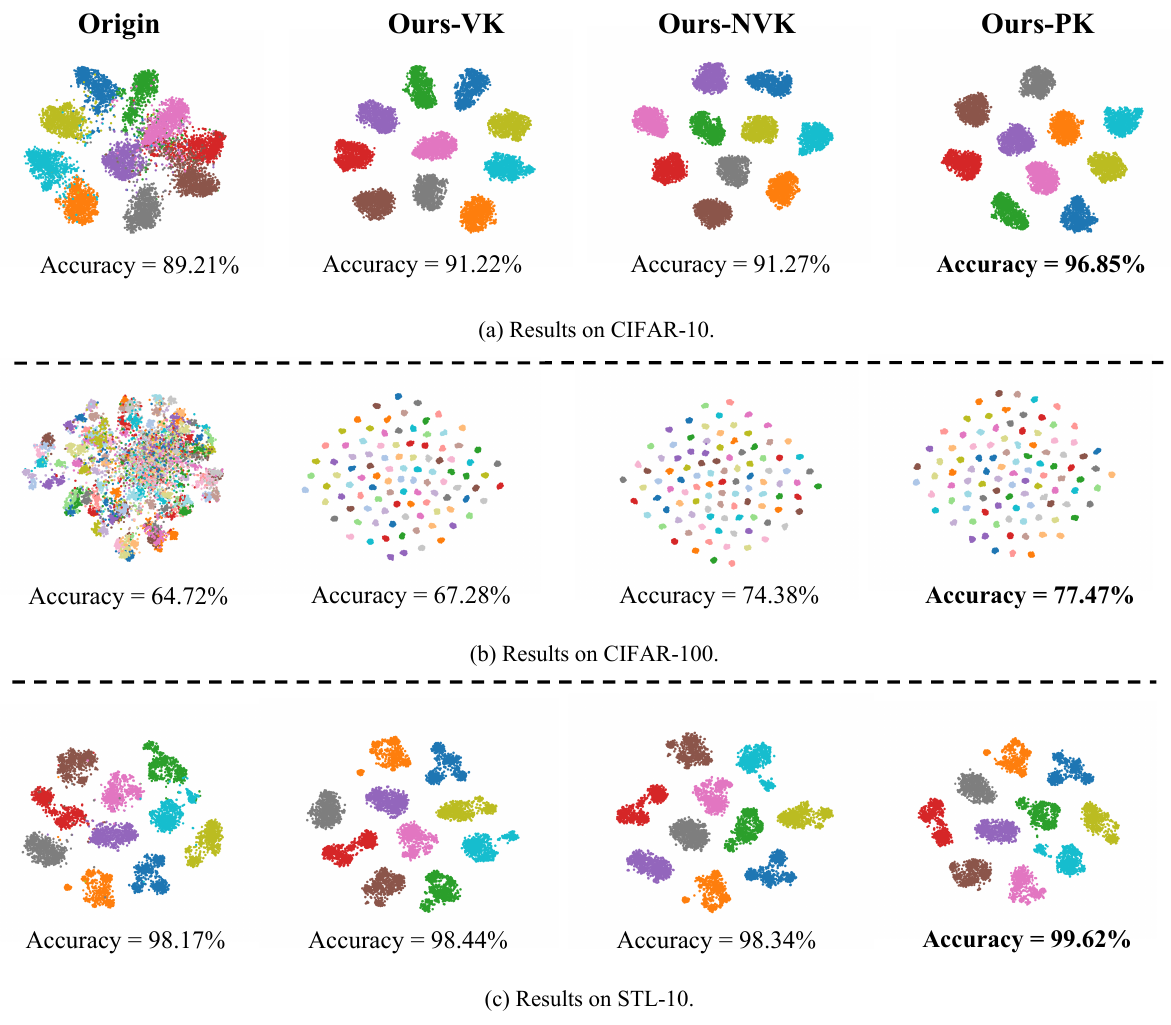}
    \caption{\textbf{t-SNE visualization of the combined vector results from the original image representation and the multi-knowledge representation. Different colors represent different categories.}}
\label{tsne}
\end{figure*}

In Table \ref{tab:comparison_method}, we clearly delineate the distinctions between our method and various others. \textit{(Extra Modules refers to incorporating other network modules for training. Soft Prompt refers to learning adaptable templates using learnable embedding vectors for downstream tasks. Image Rep represents the utilization of image representation. Fixed Template Rep represents the utilization of fixed templates from the downstream dataset, such as \textit{a photo of a [classname]}. Multi-Knowledge Rep represents the utilization of multi-knowledge representation.)} Existing approaches focus primarily on learning soft prompts and fine-tuning extra modules, lacking guidance from multi-knowledge representations. Methods centered on domain knowledge require manual adjustments of this natural language aspect, lacking an adaptive template design approach. Our proposed method, CoKnow, addresses these shortcomings. For context template optimization, we not only leverage representations of images and fixed text but also introduce additional multi-knowledge representations for context optimization, achieving outstanding results in experiments for the first time. Moreover, owing to the presence of lightweight semantic knowledge mappers in CoKnow, no additional input is required during the inference stage. The prediction results can be obtained solely by inputting the image and the trained context template.

\section{Method}
Our method is based on a pretrained vision-language model CLIP \cite{clip}, as shown in Figure \ref{coknow}. In terms of model architecture, we introduce only lightweight semantic knowledge mappers, which can generate multi-knowledge representations without additional priors. The multi-knowledge representation is crucial. To demonstrate its effectiveness, we introduce three different types of multi-knowledge representations-VK, NVK, and PK-on the CIFAR10, CIFAR100, and STL10 datasets, respectively. By combining the multi-knowledge representation with the corresponding original image representation, we obtain new representation vectors (the same approach as in Figure \ref{fig:1}(b)). In Figure \ref{zero_pre}, we present many experimental results, including various backbone visual language models, such as ResNet-50, ViT-B/32, and ViT-B/16, as well as different large language models (LLMs), including GPT-4, Qwen2 \cite{yang2024qwen2}, and Llama3.1 \cite{dubey2024llama}. These results indicate that the embedded representation of Multi-Knowledge can significantly enhance the overall predictive accuracy. However, when different LLMs are compared, the quality of Multi-Knowledge generation varies across models with different capabilities, leading to inconsistencies in predictive performance. For example, on the CIFAR10 \cite{Cifar-10}, CIFAR100 , and CUB-200-2011 \cite{WahCUB_200_2011} datasets, the multi-knowledge generated by each large language model improves the prediction accuracy of each backbone network. However, on the STL10 dataset, except for the multi-knowledge generated by GPT-4, which consistently enhances prediction accuracy, neither Qwen2 nor Llama3.1 improves accuracy across all types of multi-knowledge. At the same time, the knowledge quality generated by the powerful reasoning model GPT4-o is unstable and does not perform well on every dataset.

Furthermore, as shown in Figure 5, we perform a t-SNE visualization of the combined vector results from the original image representation and the multi-knowledge representation \textit{(Multi-Knowledge generated by GPT-4)}. On various datasets, these representations are clearly segmented into multiple clusters, corresponding to different categories. We construct a prompt learning framework guided by multi-knowledge representation. Next, we introduce the technical details and principles of our approach.

\subsection{Text Encoder Input}
This section provides a detailed introduction to the construction of inputs for the text encoder during the training phase of CoKnow. The input consists of three main parts: hand-crafted template, Multi-Knowledge, and soft prompt.

\subsubsection{Hand-Crafted Template}
Since the introduction of CLIP, the method of inserting category names into hand-crafted templates for image recognition has been widely adopted by subsequent researchers. For various downstream tasks, the hand-crafted template approach has demonstrated superior performance. We continue to leverage the hand-crafted templates proposed by previous researchers, as shown in Figure \ref{coknow}. This type of hand-crafted template is \textit{a photo of a [CLASS]}. Here, \textit{[CLASS]} represents the insertable category name.

\subsubsection{Multi-Knowledge} 
\begin{figure}
\centering
\includegraphics[width=3.5in]{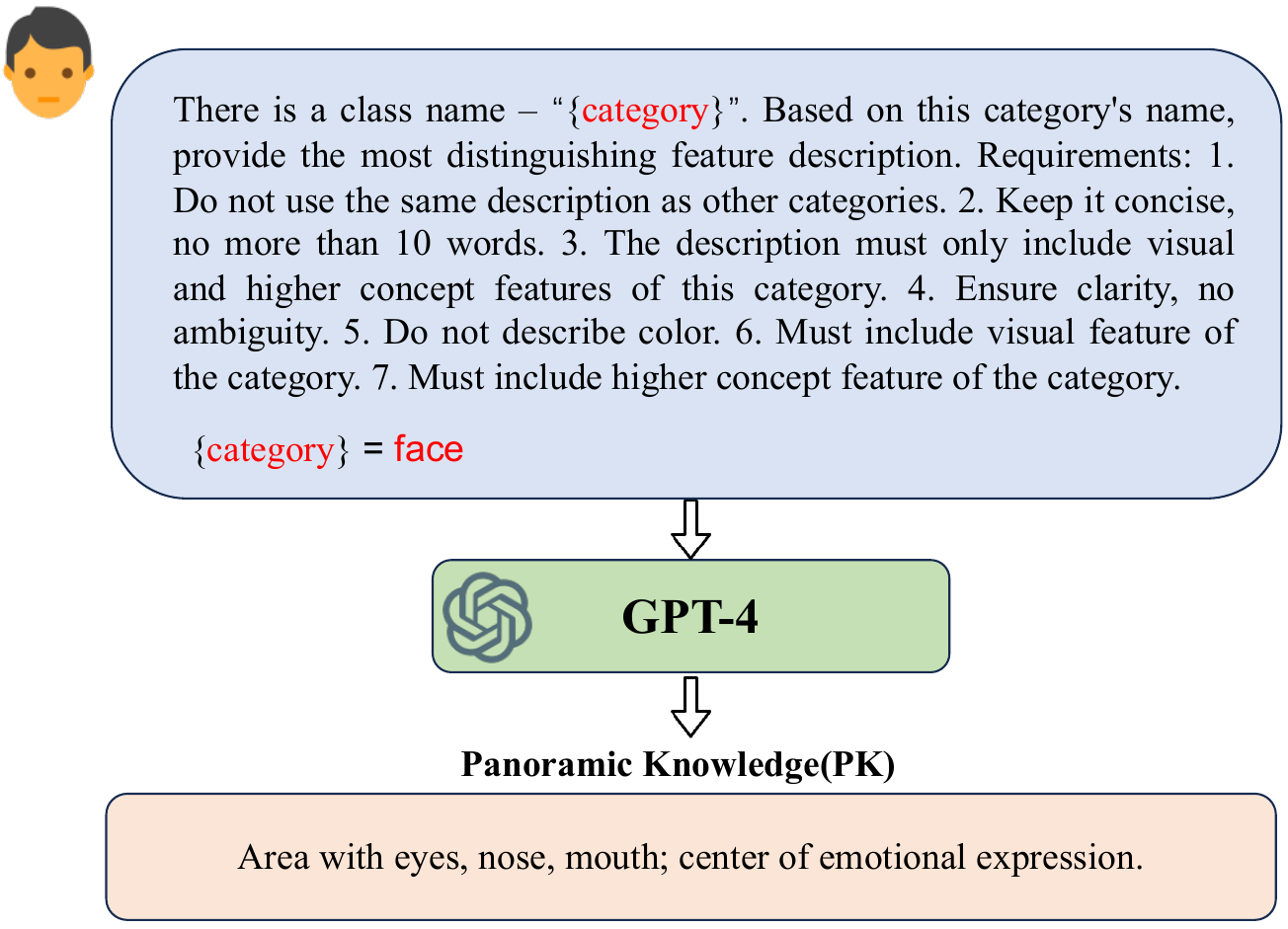}
    \caption{\textbf{The specific process of generating Multi-Knowledge. By providing the given category name along with a fixed prompt, we input these names into GPT-4 to generate a response.}}
\label{gen}
\end{figure}

We have defined three different types of Multi-Knowledge: visual knowledge (VK), nonvisual knowledge (NVK), and panoramic knwoledge (PK: VK and NVK). The first is visual knowledge (VK), which includes captions describing the image or its category. Second, and of greater importance, we introduce nonvisual knowledge (NVK) at a more abstract level beyond purely visual aspects. Third, we can combine multilevel descriptions, such as both VK and NVK, into a more comprehensive description called panoramic
knowledge (PK). We generate the Multi-Knowledge via the large language model GPT-4 by constructing different prompts. As shown in Figure \ref{coknow}, the specific method for generating the NVK is illustrated. When CLASS = \textit{airplane} is input into the prompt we constructed, we obtain the response from GPT-4: \textit{Capability to fly carrying passengers or cargo long distances.} This represents the Multi-Knowledge described from a nonvisual perspective, which is NVK. We utilize GPT-4 to construct Multi-Knowledge for downstream tasks as input to the text encoder for training. In Figure \ref{gen}, we illustrate the pipeline for generating Multi-Knowledge via GPT-4. We present a specific instance to generate a PK utilizing GPT-4. Upon providing a category \textit{face} to GPT-4 as input, the comprehensive prompt yields the following response: \textit{Area with eyes, nose, mouth; center of emotional expression.}

\subsubsection{Soft Prompt} In CoKnow, we replace text with a set of continuous vectors as learnable templates. Let $M$ be learnable vectors indexed as $\{v_1, v_2, ..., v_M\}$, where each vector corresponds to an embedding vector for a word. Let $t_i$ represent the learnable prompt constructed for the $i$-th class such that $t_i = \{v_1, v_2, ..., v_M, c_i\}$, where $c_i$ represents the embedding vector for the class name of the $i$-th category. Like hand-crafted templates do, each class name shares the same learnable prompt. As shown in Figure \ref{coknow}, the \textit{learnable context} refers to the soft prompt. We construct such soft prompts to learn effective templates adapted to downstream tasks.

\subsection{Training of CoKnow}
In this section, we introduce the complete training framework of CoKnow. The entire training framework is illustrated in Figure \ref{coknow}, where we introduce only lightweight semantic knowledge mappers as additional components to complete the entire architecture. In the text encoder, three types of templates are input: learnable context, Multi-Knowledge, and hand-crafted template. At the output end of the text encoder, three sets of representation vectors are obtained: \(T_0 = \{t_{1}^{0}, t_{2}^{0}, ..., t_{k}^{0}\}\), \(T_1 = \{t_{1}^{1}, t_{2}^{1}, ..., t_{k}^{1}\}\), and \(T_2 = \{t_{1}^{2}, t_{2}^{2}, ..., t_{k}^{2}\}\). Here, $k$ represents the number of categories. After a single image is input into the image encoder, it is duplicated into two identical visual representation vectors. These vectors then undergo processing through two semantic knowledge mappers, resulting in output vectors \(I_1\) and \(I_2\). The logits are calculated between \(I_1\) and \(T_1\), as well as between \(I_2\) and \(T_2\). Simultaneously, the vectors \(I_0\), \(I_1\), and \(I_2\) are added together via weight parameters \(\beta\) to obtain the vector \(I'\). Finally, logits are calculated between the vectors \(I'\) and \(T_0\). The specific formulas are as follows:

\begin{equation}
\text{logits}_1 = \lambda \cdot I_1 T_1^T,
\label{eq:0}
\end{equation}

\begin{equation}
\text{logits}_2 = \lambda \cdot I_2 T_2^T,
\label{eq:1}
\end{equation}

\begin{equation}
I' = I_0 \cdot \beta + I_1 \cdot \frac{1-\beta}{2} + I_2 \cdot \frac{1-\beta}{2},
\label{eq:2}
\end{equation}

\begin{equation}
\text{logits} = \lambda \cdot I' T_0^T,
\label{eq:3}
\end{equation}
\\
where \( \lambda \) refers to a scaling factor. The \( \text{logits}_1 \), \( \text{logits}_2 \), and \( \text{logits} \) are computed with the \textit{label} via cross-entropy loss. Finally, the sum of the losses from these three logits is used to update the parameters. We only train the parameters of two lightweight semantic knowledge mappers \(w_1 \) and \(w_2 \) to optimize the context templates. Our semantic knowledge mappers adopt a three-layer fully connected neural network structure consisting of an input layer, one hidden layer, and an output layer. We set the dimensionality of the hidden layer to one-fourth of the input dimension and add a ReLU activation function after the input layer. Specifically, we set the input layer dimension of the semantic knowledge mappers to correspond with the hidden size of the visual encoder (e.g., 1024). The intermediate layer dimension is set to one-fourth of the input layer dimension (e.g., 256), and the output layer is remapped to the same dimension as the input layer (1024). Additionally, we add ReLU activation functions between the input and intermediate layers, as well as between the intermediate and output layers.

\begin{figure}[!t]
\centering
\includegraphics[width=3.5in]{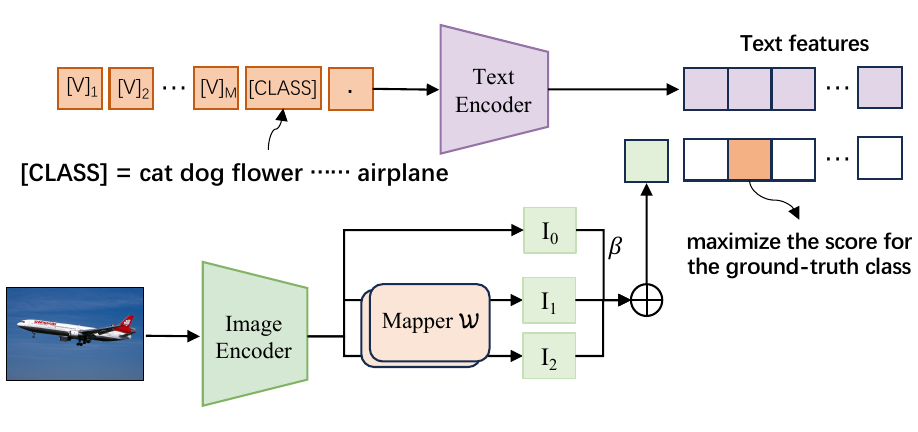}
    \caption{\textbf{The inference stage of CoKnow. The CoKnow prediction process involves automatically generating additional semantic knowledge representations when an image is input. These representations are then aggregated via $\beta$ weights to produce a new vector for prediction. During inference, our method only requires the input image and the completed training contextual template, without any additional inputs needed.}}
\label{fig4}
\end{figure}

\begin{table}
    \caption{Datasets statistics.}
    \footnotesize
    \centering
    \label{tab:datasets}
    \begin{tabular}{l r r r r}
    \toprule
    Dataset & Classes & Train & Val & Test \\
    \midrule
    ImageNet & 1,000 & 1.28M & N/A & 50,000 \\
    Caltech101 & 100 & 4,128 & 1,649 & 2,465 \\
    OxfordPets & 37 & 2,944 & 736 & 3,669 \\
    StanfordCars & 196 & 6,509 & 1,635 & 8,041 \\
    Flowers102 & 102 & 4,093 & 1,633 & 2,463 \\
    Food101\ & 101 & 50,500 & 20,200 & 30,300 \\
    FGVCAircraft & 100 & 3,334 & 3,333 & 3,333 \\
    SUN397 & 397 & 15,880 & 3,970 & 19,850 \\
    DTD & 47 & 2,820 & 1,128 & 1,692 \\
    EuroSAT & 10 & 13,500 & 5,400 & 8,100 \\
    UCF101 & 101 & 7,639 & 1,898 & 3,783 \\
    CIFAR-10 & 10 & 50,000 & N/A & 10,000 \\
    CIFAR-100 & 100 & 50,000 & N/A & 10,000 \\
    STL-10 & 10 & 5,000 & N/A & 8,000 \\
    \bottomrule
    \end{tabular}
\end{table}

\begin{table*}
    \centering
    \caption{Per-dataset results on the ResNet-50 backbone. We bold the best result for each shot and each dataset. CoKnow consistently produces the best performance across all datasets.} 
    \renewcommand{\arraystretch}{1.7}
    \resizebox{\textwidth}{!}{
    \begin{tabular}{c|c|ccccccccccc|c}
    \toprule
       &  & \multicolumn{11}{c}{\textbf{Dataset}}\\ \cmidrule(l){3-14} 
       {\textbf{Method}}  & {\textbf{Shots}} & Caltech~\cite{caltech101} & ImageNet~\cite{imagenet} & DTD~\cite{dtd} & EuroSAT~\cite{eurosat}& Aircraft~\cite{aircraft} & Food~\cite{food} & Flowers~\cite{flower} & Pets~\cite{pets} & Cars~\cite{cars} & SUN397~\cite{sun} & UCF101~\cite{ucf101} & Average \\ 

    \midrule 
    Zero-Shot CLIP  & 0     & $86.29$          & $58.18$          & $42.32$          & $37.56$          & $17.28$          & $77.31$          & $66.14$          & $85.77$          & $55.61$          & $58.52$          & $61.46$          & $58.77$          \\ \hline
               & 1     & $87.53$          & $57.15$          & $44.39$          & $50.63$          & $9.64$           & $74.32$          & $68.12$          & $85.89$          & $55.59$          & $60.29$          & $61.92$          & $59.59$          \\
                    & 2     & $87.93$          & $57.81$          & $45.15$          & $61.50$          & $18.68$          & $72.49$          & $77.51$          & $82.64$          & $58.28$          & $59.48$          & $64.09$          & $62.32$          \\ 
               {CoOp}   & 4     & $89.55$          & $59.99$          & $53.49$          & $70.18$          & $21.87$          & $73.33$          & $86.20$          & $86.70$          & $62.62$          & $63.47$          & $67.03$          & $66.77$          \\
                    & 8     & $90.21$          & $61.56$          & $59.97$          & $76.73$          & $26.13$          & $71.82$          & $91.18$          & $85.32$          & $68.43$          & $65.52$          & $71.94$          & $69.89$          \\
                    & 16    & $91.83$          & $62.95$          & $63.58$          & $83.53$          & $31.26$          & $74.67$          & $94.51$          & $87.01$          & $73.36$          & $69.26$          & $75.71$          & $73.42$          \\ \hline
                    
                    & 1     & $\textbf{89.30}$     & $\textbf{59.17}$          & $\textbf{47.93}$          & $\textbf{58.67}$         & $\textbf{18.97}$           & $\textbf{76.59}$          & $\textbf{75.83}$          & $\textbf{86.94}$          & $\textbf{57.66}$          & $\textbf{61.38}$         & $\textbf{64.69}$         & $\textbf{63.38}$          \\
                    & 2     & $\textbf{90.72}$        & $\textbf{60.07}$          & $\textbf{51.45}$          & $\textbf{66.97}$          & $\textbf{22.02}$         & $\textbf{77.35}$         & $\textbf{82.63}$          & $\textbf{87.05}$          & $\textbf{60.92}$          & $\textbf{63.98}$         & $\textbf{68.21}$          & $\textbf{66.49}$         \\ 
               CoKnow   & 4     & $\textbf{90.90}$      & $\textbf{61.00}$  &  $\textbf{58.23}$         & $\textbf{74.60}$          & $\textbf{25.67}$         & $\textbf{77.70}$          & $\textbf{88.67}$         & $\textbf{88.17}$          & $\textbf{65.67}$          & $\textbf{66.03}$         & $\textbf{71.93}$          & $\textbf{69.87}$        \\
                    & 8     & $\textbf{91.40}$          & $\textbf{62.37}$ & $\textbf{62.90}$          & $\textbf{79.83}$         & $\textbf{30.47}$         & $\textbf{78.60}$          & $\textbf{92.77}$          & $\textbf{88.90}$          & $\textbf{70.87}$          & $\textbf{68.73}$          & $\textbf{76.63}$          & $\textbf{73.04}$          \\
                    & 16    & $\textbf{92.83}$       & $\textbf{63.83}$   & $\textbf{67.20}$       & $\textbf{84.77}$         & $\textbf{37.57}$         & $\textbf{78.83}$         & $\textbf{95.33}$          & $\textbf{89.13}$          & $\textbf{76.67}$          & $\textbf{71.07}$          & $\textbf{79.77}$          & $\textbf{76.09}$          \\
                    
                    \midrule 
    \end{tabular}}
    \label{tab:per_dataset}
\end{table*}

\subsection{Inference of CoKnow}
In this section, we introduce the complete inference framework of CoKnow. As illustrated in Figure \ref{fig4}, our method is very concise during the inference stage, requiring only the input of the image and the learned context template, without any additional input. Assuming that there are $K$ categories in total, where $w_i$ represents the output of the text encoder when the category is $i$, $x_0$ represents the output of the image encoder for the given image $I$, and $x_1$ and $x_2$ represent the outputs of the semantic knowledge mappers $W$ for the given image $I$; the probability of it belonging to category $i$ is as follows:

\begin{equation}
x = \beta \cdot x_0 + (\frac{1-\beta }{2} )\cdot x_1 + (\frac{1-\beta }{2} )\cdot x_2,
\label{eq:4}
\end{equation}

\begin{equation}
p(y = i | I) = \frac{\exp(\text{sim}(w_i, x) / \tau)}{\sum_{k=1}^{K} \exp(\text{sim}(w_k, x) / \tau)},
\label{eq:5}
\end{equation}
\\
where \(\text{sim}(w_i, x)\) denotes the cosine similarity between \(w_i\) and \(x\), \(\beta\) represents the weight parameter, and \(\tau\) is a temperature parameter.

\section{Experiments}

\begin{table}[t]
    \centering
    \caption{Comparison with previous methods. We used ResNet50 or ViT-B/16 as the backbone to compare the average top-1 accuracy across 11 test sets under the condition of few-shot learning with five different sample sizes. The best results are highlighted in bold, and the second-best results are underlined.}
    \setlength{\tabcolsep}{3pt} 
    \renewcommand{\arraystretch}{1.4}
    \footnotesize
    \begin{tabular}{l|c|ccccc}
        \toprule
        \quad \quad \textbf{Method} & \textbf{Ref}. & \multicolumn{5}{c}{\textbf{Number of Shots}} \\ 
        & & 1 & 2 & 4 & 8 & 16 \\ 
        \midrule
        \multicolumn{7}{l}{\textit{ResNet-50}} \\
        \midrule
        CoOp & IJCV'22 & 59.59 & 62.32 & 66.77 & 69.89 & 73.42 \\
        Wise-FT & CVPR'22 & 59.09 & 61.80 & 65.29 & 68.43 & 71.64 \\
        ProGrad & ICCV'23 & 62.61 & 64.90 & 68.45 & 71.41 & 73.96 \\
        Tip-Adapter-F & ECCV'22 & 63.25 & 65.93 & 68.98 & \underline{72.15} & \underline{75.10} \\
        Cross-Modal Wise-FT & CVPR'23 & \textbf{63.76} & \underline{66.40} & 68.95 & 71.73 & 74.08 \\
        \textbf{CoKnow} & -- & \underline{63.38} & \textbf{66.49} & \textbf{69.87} & \textbf{73.04} & \textbf{76.09} \\
        \midrule
        \multicolumn{7}{l}{\textit{ViT-B/16}} \\
        \midrule
        CoCoOp & CVPR'22 & 69.10 & 70.38 & 72.32 & 76.20 & 78.43 \\
        Tip-Adapter & ECCV'22 & \underline{69.81} & \underline{71.56} & \underline{74.18} & 75.17 & 77.39 \\
        Cross-Modal WiSE-FT & CVPR'23 & \textbf{70.08} & 71.13 & 72.97 & \underline{76.96} & 79.45 \\
        KgCoOp & CVPR'23 & 69.30 & 70.68 & 72.51 & 76.55 & 78.71 \\
        \textbf{CoKnow} & - & 69.44 & \textbf{72.67} & \textbf{74.93} & \textbf{77.80} & \textbf{80.78} \\
        \bottomrule
    \end{tabular}
    \label{tab:comparison_sota}
\end{table}

\begin{table}[t]
    \centering
    \caption{Comparison on the distribution shift. In the context of distribution shift, we have conducted comparisons with previous works. We employed ViT-B/16 for training on the ImageNet dataset, followed by zero-shot testing on the ImageNet-V2 datasets.}
    \label{tab:ood}
    \renewcommand{\arraystretch}{1.5} 
    \setlength{\tabcolsep}{3pt} 
    \scalebox{1.00}{ 
        \begin{tabular}{l|c|cc}
        \toprule
        Method & ImageNet & ImageNet-V2 \\
        \midrule
        Zero-Shot CLIP & 66.7 & 60.8 \\
        Linear Probe & 65.9 & 56.3 \\
        CoOp\cite{coop} & 71.7 & 64.6 \\
        CoCoOp\cite{cocoop} & 71.0 & 64.1 \\
        KgCoOp\cite{kg} & 71.2 & 64.1 \\
        \textbf{CoKnow} & \textbf{72.07} & \textbf{64.7} \\
        \bottomrule
        \end{tabular}}
\end{table}

\subsection{Experimental Settings}

\subsubsection{Datasets} We conducted experiments on 11 publicly available datasets, encompassing a wide range of scenes and rich content, including ImageNet \cite{imagenet}, Caltech \cite{caltech101}, Oxford-Pets \cite{pets}, Flowers \cite{flower}, Food101 \cite{food}, Stanford Cars \cite{cars}, FGVCAircraft \cite{aircraft}, EuroSAT \cite{eurosat}, UCF101 \cite{ucf101}, DTD \cite{dtd}, and SUN397 \cite{sun}. In Table 6, we present the statistical information of the datasets. In strict adherence to the few-shot evaluation protocol outlined in CoOp \cite{coop}, we utilized 1, 2, 4, 8, and 16 shots for training and validated the results on the complete test sets.

\subsubsection{Implementation Details} To ensure a fair comparison, we strictly adhered to the CoOp protocol \cite{coop}, employing ResNet-50 or ViT-B/16 as the backbone architecture for the CLIP image encoder. Additionally, we evaluated the performance using ViT-B/16 as the vision encoder. The language model used maintains the same temperature parameter across each dataset, set to the default value. Our experiments were conducted with a batch size of 32 and epoch sizes ranging from 10 to 400. Learning rates of 0.001 and 0.002 were utilized alongside the SGD optimizer and a cosine annealing strategy. The training process was executed on 6 NVIDIA GeForce RTX 3090 GPUs. We utilize three types of Multi-Knowledge through GPT-4 \textit{(GPT-4 refers to the large language model endpoint "gpt-4" developed by OpenAI.)} These include visual knowledge (VK), nonvisual knowledge (NVK), and panoramic knowledge (PK: VK and NVK). In the experiments of \textit{Few-shot Learning} and \textit{Distribution Shift}, we set the \(\beta\) value at 0.6, which was determined to yield the optimal experimental results. In the ablation studies, we accordingly explored the impact of varying \(\beta\) values on the experiments, and the specific \(\beta\) values employed can be found in the corresponding ablation study sections.

\subsection{Performance}
To assess the effectiveness of our approach, we conducted fair comparisons with a range of prior works, including CoOp \cite{coop}, CoCoOp \cite{cocoop}, Wise-FT \cite{WiSE-FT}, ProGrad \cite{ProGrad}, Tip-Adapter-F \cite{Tip-adapter}, Cross-Modal Wise-FT \cite{Cross-modal}, and Linear-Probe CLIP \cite{clip}.

\begin{figure*}[t]
    \centering
    \includegraphics[width=0.90\textwidth]{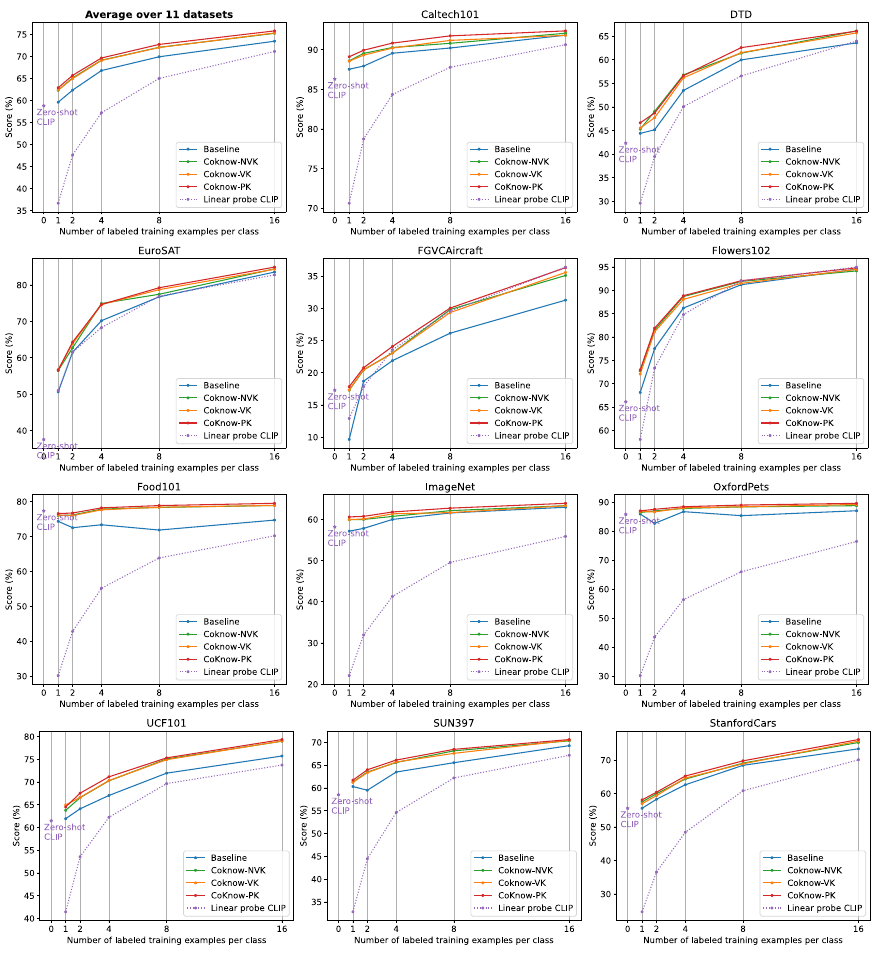}
    \caption{\textbf{Ablation study on inputting different Multi-Knowledge. At the text input end, in addition to the hand-crafted template, we input different muultiknowledge, including NVK, VK, and PK, as target multi-knowledge representations for learning. \textit{The \textit{Baseline} represents direct prompt learning via CoOp.}}}
    \label{abl_k}
\end{figure*}

\begin{table}[t]
    \centering
    \caption{Multi-Knowledge Ablation Experiment ($\beta=0.6$).}
    \label{tab:abl_0.6}
    \renewcommand{\arraystretch}{1.7} 
    \footnotesize
    \setlength{\tabcolsep}{5pt} 
    \scalebox{1.0}{ 
        \begin{tabular}{ccc|ccccc}
        \toprule
        \multicolumn{3}{c|}{\textbf{Multi-knowledge}} & \multicolumn{5}{c}{\textbf{Number of Shots}} \\
        \midrule
        \quad \textbf{NVK} & \textbf{VK} & \textbf{PK} & 1 & 2 & 4 & 8 & 16 \\
        \midrule
        \centering -- & \centering -- & \quad --\hspace*{\fill} & 59.59 & 62.32 & 66.77 & 69.89 & 73.42 \\
        \quad \centering\checkmark & \centering & \hspace*{\fill} & \underline{63.10} & \underline{66.12} & \underline{69.73} & 72.75 & \underline{75.68} \\
        \centering & \centering\checkmark & \hspace*{\fill} & 63.01 & 66.03 & 69.35 & \underline{72.76} & 75.65 \\
        \centering & \centering & \quad \checkmark\hspace*{\fill} & \textbf{63.38} & \textbf{66.49} & \textbf{69.87} & \textbf{73.04} & \textbf{76.09} \\
        \bottomrule
        \end{tabular}
    }
\end{table}

\subsubsection{Few-shot Learning} In Table \ref{tab:per_dataset}, we present a comparison of the average top-1 accuracy rates of CoKnow with CoOp across each dataset. This intuitively detailed display highlights the performance difference between CoKnow and prompt learning methods solely utilizing image representations, with CoOp serving as our baseline. Notably, for every accuracy metric, we have demonstrated superior performance. In Table \ref{tab:comparison_sota}, we compared our approach with a greater variety of methods than before. Our method surpasses a range of previous approaches from 2 shots to 16 shots. Notably, in the 1-shot scenario, CoKnow outperforms both CoOp and Wise-FT in terms of the 2 shot results. With 4 shots, CoKnow's results approach those of CoOp with 8 shots and surpass the results of Wise-FT. In the 8-shot scenario, the CoKnow results are comparable to those of CoOp and even surpass those of Wise-FT. These experimental results demonstrate the effectiveness of CoKnow in learning from limited data.

\subsubsection{Distribution Shift} We further evaluated the robustness of CoKnow under out-of-distribution (OOD) conditions. We conducted prompt learning on ImageNet and tested it on ImageNetV2 \cite{V2}. The results are presented in Table \ref{tab:ood}. The experimental results indicate that the method effectively generalizes from ImageNet to out-of-distribution datasets, highlighting its potential in handling distribution shifts.

\subsubsection{Inference Overhead} We present the costs of training and inference for CoKnow on datasets such as Caltech101, Pets, and Cars in Table \ref{overhead}. The table presents the results when the batch size is 32 and the learnable prompt length is 16. The lightweight framework of CoKnow effectively reduces the inference overhead, making it easier to apply to downstream tasks. Specifically, the semantic knowledge mapper in the CoKnow architecture allows us to avoid loading large amounts of Multi-Knowledge during the inference phase, thereby effectively reducing the inference overhead. For example, the memory usage on the StanfordCars dataset decreases from 4002 MB to 1434 MB.

\begin{table}[h]
    \centering
    \caption{Overhead during training and inference.}
    \label{overhead}
    \begin{tabular}{ccc}
        \toprule
        \textbf{Datasets} & \textbf{Training (MB)} & \textbf{Inference (MB)} $\downarrow$ \\ 
        \midrule
        OxfordPets        & 1370                  & 932                                         \\ 
        Caltech101        & 2488                  & 988                                         \\ 
        StanfordCars      & 4002                  & 1434                                        \\ 
        \bottomrule
    \end{tabular}
\end{table}

\begin{figure}[t]
\centering
\hspace{-5.0mm}
\includegraphics[width=3.2in]{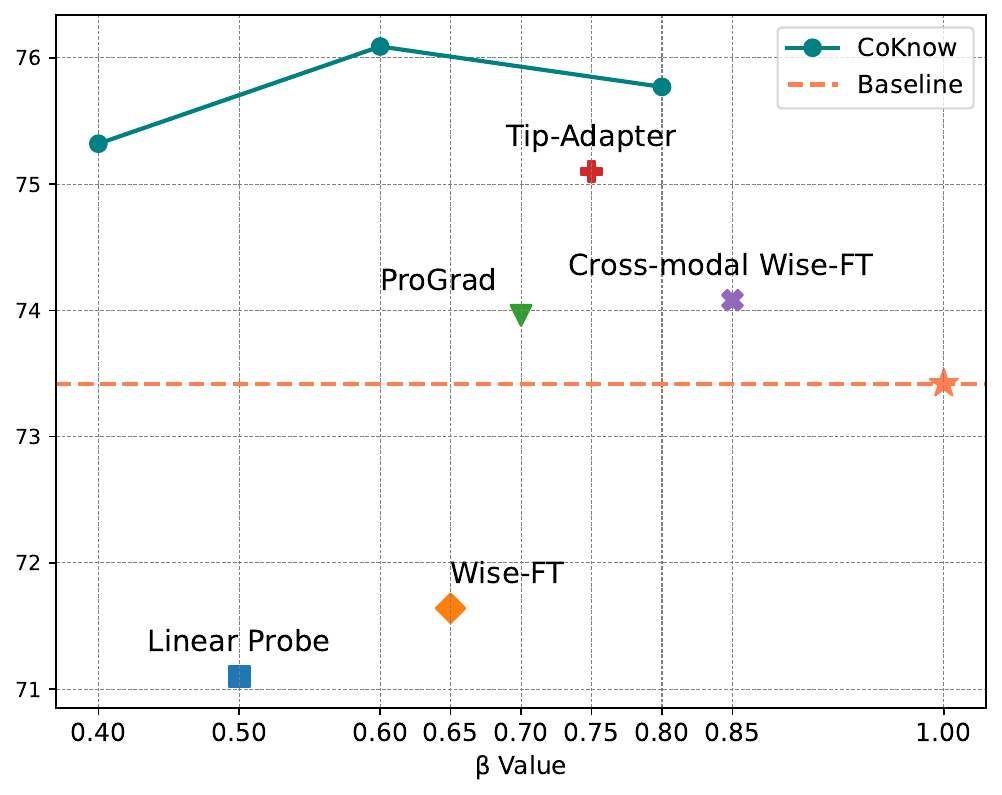}
    \caption{\textbf{Average prediction results for different $\beta$ values. The accuracy of a series of previous methods is directly marked on the graph.
}}
\label{beta}
\end{figure}

\subsection{Ablation Study}

\subsubsection{Different Multi-Knowledge} We utilize VK, NVK, and PK, three different types of knowledge descriptions, to further explore the impact of different Multi-Knowledge on prompt learning results when the same hand-crafted template is input. Setting \(\beta\) at 0.8 and optimizing the context length to 16, we detail the average effects of few-shot learning across 11 datasets in Figure \ref{abl_k}. Overall, descriptions of varying Multi-Knowledge were able to further optimize prompt learning, achieving commendable levels of performance. The overall prediction accuracy of PK is the best, indicating that a global representation incorporating knowledge from multiple perspectives is more effective. We present the impact of different Multi-Knowledge scenarios with \(\beta\) set to 0.6 in Table \ref{tab:abl_0.6} for further demonstration.

\begin{table*}[t]
    \centering
    \caption{Ablation Study of Mapper. CoKnow-I represents the re-mapping of the original CLIP's image representations in CoKnow using a semantic knowledge mapper.}  
    \renewcommand{\arraystretch}{1.5}
    \resizebox{\textwidth}{!}{
    \begin{tabular}{c|c|ccccccccccc|c}
    \toprule
       &  & \multicolumn{11}{c}{\textbf{Dataset}}\\ \cmidrule(l){3-14} 
       {\textbf{Method}}  & {\textbf{Shots}} & Caltech~\cite{caltech101} & ImageNet~\cite{imagenet} & DTD~\cite{dtd} & EuroSAT~\cite{eurosat}& Aircraft~\cite{aircraft} & Food~\cite{food} & Flowers~\cite{flower} & Pets~\cite{pets} & Cars~\cite{cars} & SUN397~\cite{sun} & UCF101~\cite{ucf101} & Average \\ 

    \midrule 
            \multirow{2}{*}{CoKnow-I}   & 1     & $47.23$          & $27.40$          & $27.67$          & $53.07$          & $13.13$           & $13.33$          & $56.27$          & $10.13$          & $32.13$          & $26.03$          & $22.43$          & $29.89$          \\
                   & 16     & $91.13$          & $60.52$          & $63.77$          & $83.40$          & $35.13$          & $68.97$          & $94.10$          & $78.3$          & $72.57$          & $65.20$          & $74.70$          & $71.10$          \\ \hline
                    
               \multirow{2}{*}{\textbf{CoKnow}}   & 1     & $\textbf{89.30}$     & $\textbf{59.17}$          & $\textbf{47.93}$          & $\textbf{58.67}$         & $\textbf{18.97}$           & $\textbf{76.59}$          & $\textbf{75.83}$          & $\textbf{86.94}$          & $\textbf{57.66}$          & $\textbf{61.38}$         & $\textbf{64.69}$         & $\textbf{63.38}$          \\
                    & 16    & $\textbf{92.83}$       & $\textbf{63.83}$   & $\textbf{67.20}$       & $\textbf{84.77}$         & $\textbf{37.57}$         & $\textbf{78.83}$         & $\textbf{95.33}$          & $\textbf{89.13}$          & $\textbf{76.67}$          & $\textbf{71.07}$          & $\textbf{79.77}$          & $\textbf{76.09}$          \\
                    \midrule 
    \end{tabular}}
    \label{tab:mapper}
\end{table*}

\begin{table}[t]
    \centering
    \caption{Ablation Study on Different Classname Positions. \textit{End} represents the category name located at the end of the learnable vectors, whereas \textit{Middle} represents the category name located in the middle of the learnable vectors.}
    \label{tab:abl_2}
    \renewcommand{\arraystretch}{1.4} 
    \setlength{\tabcolsep}{4pt} 
    \scalebox{1.0}{ 
        \begin{tabular}{cc|ccccc}
        \toprule
        \multicolumn{2}{c|}{\textbf{Position of Classnames}} & \multicolumn{5}{c}{\textbf{Number of Shots}} \\
        \midrule
        \quad \textbf{End} & \textbf{Middle} & 1 & 2 & 4 & 8 & 16 \\
        \midrule
        \cmark & & \textbf{63.31} & \textbf{66.00} & \textbf{69.87} & 73.04 & \textbf{76.09} \\
        \addlinespace[0.5em]
        & \cmark & 61.91 & 65.63 & 69.58 & \textbf{73.11} & 76.01 \\
        \bottomrule
        \end{tabular}
    }
\end{table}

\begin{table}[h]
\centering
\caption{Average Prediction Results for Different Context Lengths. This represents the lengths of learnable vectors of different sizes.}
\label{tab:len}
\renewcommand{\arraystretch}{1.4}
\begin{tabular}{cc}
\hline
\textbf{Context Length}  & \textbf{Accuracy(\%)} \\ \hline
8                  & 76.02                \\
16                 & \textbf{76.09}                \\
32                 & 75.96                \\ \hline
\end{tabular}
\end{table}

\subsubsection{Different $\beta$ weight values.} We investigated the impact of different $\beta$ weights on the model's outcomes by setting the $\beta$ values to 0.4, 0.6, and 0.8, with PK as the input and the optimization context length set to 16. Figure \ref{beta} presents a comparison of the average predictions across 11 datasets under the condition of 16 shots. Simultaneously, the effects of previous methods are directly demonstrated in the graph. Different values of $\beta$ have a certain impact on the results, but they all outperform a series of previous excellent methods. Our experiments offer guidance for adjusting the $\beta$ value.

\subsubsection{Ablation Study of Mapper} This experiment explores the effectiveness of using a semantic knowledge mapper to map the image representations of the original CLIP (denoted as $I_0$ in Figure \ref{coknow}) and investigates the importance of the original CLIP's image representations for prompt learning. The $\beta$ value is set to 0.6, the input is set to PK, and the optimized context length is set to 16. We incorporate a semantic knowledge mapper before $I_0$ in Figure \ref{coknow}, and the new structure is referred to as \textit{CoKnow-I}. As shown in Table \ref{tab:mapper}, when the training sample size is 1 shot, CoKnow-I achieves very low Top-1 accuracy across 11 datasets. When the training sample size is increased to 16 shots, the Top-1 accuracy across these 11 datasets remains lower than the predicted results of CoKnow. The experimental results indicate that the image representations of the original CLIP have a significant effect on prompt learning, especially when the training sample size is 1 shot. Additionally, it is not advisable to use a semantic knowledge mapper to map the image representations of the original CLIP.

\subsubsection{Position of Classname Varies} In the context of optimization, the position of the \textit{classname} can be either at the end or in the middle \cite{coop}. Setting $\beta = 0.6$, the context length is optimized to 16, and PK is used as the input. We explored the impact of different \textit{classname} positions on the experimental results across 11 datasets. The outcomes are presented in Table \ref{tab:abl_2}. The overall performance is better when the \textit{classname} position is at the end rather than in the middle.

\subsubsection{Different Context Lengths} We set the context lengths to 8, 16, and 32, with PK as the input and the $\beta$ value set to 0.6. We evaluated the mean comparison after training with 16 shots on 11 datasets to explore the impact of different context lengths on the results. The results are shown in Table \ref{tab:len}. Our method demonstrates robust performance across varying context lengths, with minimal perturbation to the outcomes attributable to changes in context length.

\subsubsection{Longer Embeddings} To demonstrate that our method improves prediction performance through different types of multi-knowledge rather than by using longer embeddings, we employ three independent weight heads for contrastive learning. We then extend the length of the embeddings by concatenation for prediction. We use PK with ResNet50 as the backbone network, setting the $\beta$ value to 0.6. Three independent heads expand the embedding dimension from the original 1024 to 3072, represented by \textit{CoKnow-LE}. The experimental results are presented in Table \ref{tab:le}. The experimental results show that longer embeddings do not effectively improve prediction performance.

\begin{table}[t]
    \centering
    \caption{Ablation Study of Longer Embeddings.}  
    \renewcommand{\arraystretch}{1.5}
    \resizebox{\columnwidth}{!}{
    \begin{tabular}{c|c|ccccc|c}
    \toprule
       &  & \multicolumn{5}{c}{\textbf{Dataset}} &  \\ \cmidrule(l){3-8} 
       {\textbf{Method}}  & {\textbf{Shots}} & Caltech~\cite{caltech101} & DTD~\cite{dtd} & EuroSAT~\cite{eurosat} & Aircraft~\cite{aircraft} & Flowers~\cite{flower} & \textbf{Average}\\

    \midrule 
            \multirow{5}{*}{CoKnow-LE}   & 1     & $69.47$          & $31.07$          & $51.63$          & $15.07$           & $44.13$          & $42.27$          \\
                   & 2     & $76.13$          & $41.00$          & $55.93$          & $18.07$          & $56.30$          & $49.49$          \\
                   & 4     & $83.67$          & $52.63$          & $60.53$          & $21.93$          & $55.97$          & $54.95$          \\
                   & 8     & $88.10$          & $60.73$          & $69.60$          & $27.87$          & $88.20$          & $66.90$          \\
                   & 16    & $90.90$          & $66.37$          & $77.50$          & $35.20$          & $93.57$          & $72.71$          \\ \midrule
                    
               \multirow{5}{*}{\textbf{CoKnow(ours)}}   & 1     & $\textbf{89.30}$     & $\textbf{47.93}$          & $\textbf{58.67}$         & $\textbf{18.97}$           & $\textbf{75.83}$          & $\textbf{58.14}$          \\
                    & 2     & $\textbf{90.72}$       & $\textbf{51.45}$       & $\textbf{66.97}$         & $\textbf{22.02}$         & $\textbf{82.63}$          & $\textbf{62.76}$          \\
                    & 4     & $\textbf{90.90}$       & $\textbf{58.23}$       & $\textbf{74.60}$         & $\textbf{25.67}$         & $\textbf{88.67}$          & $\textbf{67.61}$          \\
                    & 8     & $\textbf{91.40}$       & $\textbf{62.90}$       & $\textbf{79.83}$         & $\textbf{30.47}$         & $\textbf{92.77}$          & $\textbf{71.47}$          \\
                    & 16    & $\textbf{92.83}$       & $\textbf{67.20}$       & $\textbf{84.77}$         & $\textbf{37.57}$         & $\textbf{95.33}$          & $\textbf{93.10}$          \\
                    \midrule 
    \end{tabular}}
    \label{tab:le}
\end{table}

\section{Limitations}

This paper introduces Multi-Knowledge and effectively enhances the generalization performance of VLMs in downstream tasks. However, the ability of LLMs to generate the highest-quality Multi-Knowledge for maximizing the predictive power of VLMs has not yet been fully explored. As illustrated in Figure \ref{zero_pre}, using different LLMs requires further consideration of prompt strategies tailored to each model to fully leverage its potential. In the CoKnow framework proposed in this paper, the mapper layer employs a simple fully connected layer structure. As shown in Table \ref{tab:comparison_sota}, CoKnow does not achieve the highest performance metrics in the 1-shot setting, indicating that the generalization performance of this mapper structure needs improvement when the sample size is minimal. More advanced designs are needed in the future to address this issue.

\section{Conclusion}
In this paper, we gain insights into the importance of multi-knowledge representation for VLMs. Existing technologies have overlooked the method of context template learning, which incorporates domain knowledge. Such oversight will restrict the ability of VLMs to learn multi-knowledge representations during the pretraining phase, further limiting its performance in downstream tasks. To address this issue, we propose context optimization with multi-knowledge representation (CoKnow). We trained lightweight semantic knowledge mappers capable of generating multi-knowledge representations without requiring additional inputs while also constructing a knowledge-guided prompt learning framework. We extensively evaluated it on multiple benchmark tests, and the results indicate that the proposed CoKnow is an effective method for prompt learning.

{
\bibliographystyle{ieeetr}
\bibliography{ref}
}

\end{document}